\documentclass{article}
\usepackage{ijcai25}

\usepackage{times}
\usepackage{soul}
\usepackage{url}
\usepackage[hidelinks]{hyperref}
\usepackage[utf8]{inputenc}
\usepackage[small]{caption}
\usepackage{graphicx}
\usepackage{amsmath}
\usepackage{amsthm}
\usepackage{booktabs}
\usepackage{algorithm}
\usepackage{algorithmic}
\usepackage[switch]{lineno}

\usepackage{hyperref}       
\usepackage{url}
\usepackage{color}
\usepackage{makecell}
\usepackage[caption=false]{subfig}
\usepackage{bm}
\usepackage{array} 

\definecolor{darkgreen}{rgb}{.2, 0.5, 0.2}

\let\mc\mathcal

\newtheorem{definition}{Definition}

\newtheorem{example}{Example}

\title{Conversation Games and a Strategic View of the Turing Test}

\author{
    Kaveh Aryan \\
    King's College London \\
     \href{mailto:kaveh.aryan@kcl.ac.uk}{kaveh.aryan@kcl.ac.uk}
}

\begin{document}

\maketitle

\begin{abstract}
Although many game-theoretic models replicate real interactions that often rely on natural language, explicit study of games where language is central to strategic interaction remains limited. This paper introduces the \emph{conversation game}, a multi-stage, extensive-form game based on linguistic strategic interaction. We focus on a subset of the games, called verdict games. In a verdict game, two players alternate to contribute to a conversation, which is evaluated at each stage by a non-strategic judge who may render a conclusive binary verdict,  or a decision to continue the dialogue. The game ends once a limit is reached or a verdict is given. We show many familiar processes, such as interrogation or a court process fall under this category. We also, show that the Turing test is an instance of verdict game, and discuss the significance of a strategic view of the Turing test in the age of advanced AI deception. We show the practical relevance of the proposed concepts by simulation experiments, and show that a strategic agent outperforms a naive agent by a high margin.
\end{abstract}

\section{Introduction}
Language functions not only as a channel for communication but also as a strategic instrument, whether in legal disputes, everyday conversations, or high-stakes negotiations. However, it has as not traditionally been studied this way. In this paper, we focus on a class of dialogue-driven interactions that we call \emph{conversation games}. Our goal is to provide a formal framework that captures the strategic essence of these interactions, allowing us to analyse how participants exchange information and manipulate discourse to achieve specific objectives. After motivating examples, we introduce a general definition of conversation games and then specialize to the subset of games, called \emph{verdict games}, in which a conversation leads to an external judgment, such as a court ruling or a verdict of authenticity, while players have private types and distinct incentives. We show that the Turing test \cite{oppy_turing_2021} is an instance of the verdict game and discuss its strategic aspects. Furthermore, our simulation experiments demonstrate that strategic agents outperform naive ones and Demonstrates the relevance of the proposed concept. The rest of this paper is structured as follows: Sections \ref{sec:conversation-game} and \ref{sec:verdict-games} introduce the notions of \emph{conversation games} and \emph{verdict games}, highlighting their significance through various examples. Section \ref{sec:experiments} discusses our experiments, Section \ref{sec:related-works} reviews related works, and Section \ref{sec:conclusion} concludes the paper.

\section{Conversation Game} \label{sec:conversation-game}
Language is a ubiquitous medium for strategic interactions in real-world settings, from negotiations and debates to customer support and collaborative problem-solving. Unlike traditional games, where moves are tangible actions, conversation games involve contributions to an evolving dialogue. Understanding these dynamics enables researchers to model and predict behaviour in domains like diplomacy, AI-human interactions, and social decision-making.

\begin{example} \label{ex:everyday-life}
We begin by highlighting how people use language strategically, leveraging nuances, ambiguity, and calculated phrasing to influence outcomes and manage relationships. Whether it is in workplace (``To deliver the best results, it might be helpful to revisit the timeline?''), friendship (``I’ve heard journaling can be really helpful.''), or dating (``That new exhibit sounds interesting!''), language is used strategically as tool for navigating social interactions and achieving goals.
\end{example}

\begin{example} \label{ex:court}
\textbf{(Court)} One of the most notable examples of a conversation game is a courtroom trial. Here, the players are the prosecution and the defense, each strategically interacting to achieve their respective goals (conviction or exoneration). Although physical or forensic evidence may be presented, the dialogue surrounding these pieces of evidence often drives the judge or jury’s perception. Each utterance aims to persuade the fact-finders and shape their interpretation of the evidence. 
\end{example}

Conversation games can involve hidden roles and Bayesian reasoning, as the following couple of examples illustrate:
\begin{example} \label{ex:interrogation}
\textbf{(Interrogation)} In an interrogation setting, the key players are the interrogator and the suspect. In many legal contexts, the ultimate goal is not merely to gather intelligence but to do so in a manner admissible and convincing in court. Consequently, the interrogator must guide the dialogue to elicit information in a form that will be deemed credible and legally defensible, while the suspect may strategically reveal or conceal details to influence legal outcomes.
\end{example}

\begin{example} \label{ex:turing-test}
\textbf{(Turing test)} Another notable example is the Turing test \cite{oppy_turing_2021}. Here, the players are a human interrogator and a witness who may be either human or machine. The interrogator updates their beliefs (in a Bayesian fashion) about the witness's identity based on the responses received. The witness, in turn, strategically produces language that minimises the chances of being identified as .
\end{example}

It should be noted that the utility of players is not necessarily determined by the conversation itself, but rather by some external effect or outcome of the conversation:

\begin{example}
\textbf{(Psychoanalysis)} In a psychoanalysis or cognitive behavioral therapy (CBT) session, the therapist and the patient engage in a conversation aimed at improving the patient's mental well-being. Although the dialogue is central to the therapeutic process, the eventual measure of success lies in the patient's external life improvements or changes in mental health, rather than the conversation itself.
\end{example}

\begin{example}
\textbf{(Teaching)} In a classroom setting, a teacher and students engage in conversational exchanges. The teacher’s ultimate objective is effective learning as evidenced by students’ grasp of the material, rather than simply guiding a smooth dialogue. 
\end{example}

One can readily identify other examples that are subsumed under the overarching concept of the conversation game, such as diplomacy, bargaining, negotiation, debate, and persuasion. A formal definition of the conversation game is as follow: 

\begin{definition}
\textbf{Conversation Game.}
Formally, a conversation game is a multistage, extensive-form game with the following characteristics: 
\begin{enumerate}
    \item \emph{Players and roles.} There can be multiple players, each potentially holding a private type.
    \item \emph{Actions as utterances.} At each stage, a player's action is an utterance, which contributes to the evolving dialogue.
    \item \emph{Information structure.} Players may have private information about their type or objectives, and they update their beliefs based on the content of the conversation.
    \item \emph{Payoffs.} The utility of each player is determined by the ultimate outcome, which may be influenced but not fully determined by the conversation's contents.
\end{enumerate}
\end{definition}

In this paper, we primarily focus on a type of conversation game illustrated by Examples \ref{ex:court}, \ref{ex:interrogation}, and \ref{ex:turing-test}. These are games where players guide the conversation so that a non-strategic judge (or evaluator) can make a binary judgment solely based on the dialogue's content. The utility of the players depends on their private types and on the outcome of this external judgment. We refer to this subset of conversation games as \emph{verdict games}, and they will be formalized in detail in the following section.

\section{Verdict Games} \label{sec:verdict-games}
A \emph{verdict game} is a conversation game in which players' utilities are determined by the outcome of a binary sequence classifier on the conversation history. 

\subsection{Problem Setup}
\begin{figure}[!t]
  \centering
  \includegraphics[width=0.5\textwidth]{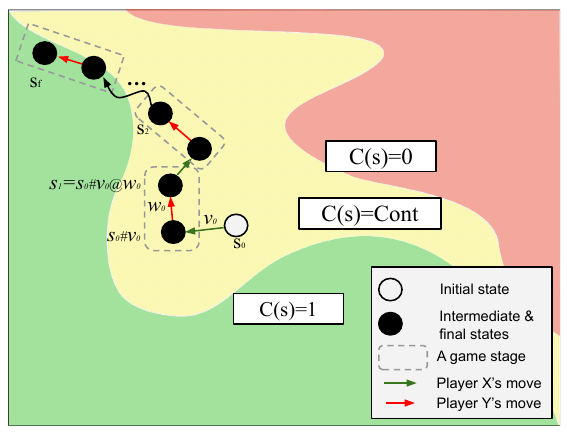}
  \caption{An illustration of the verdict game.}
  \label{fig:verdict-game}
\end{figure}

The \emph{verdict game} (Fig. \ref{fig:verdict-game}) game needs to be general to accommodate different games and player goals. The game involves two players, player \(X\) (she/her/her) and player \(Y\) (he/his/him). Each player has a type drawn from their respective set, \(T_X\) for \(X\) and \(T_Y\) for \(Y\), which might or might not be revealed to the other. The game starts at state \(s_0\), which is a string (possibly empty) from an alphabet \(\Sigma\). The game is played in multiple stages. Each stage starts with the \(X\) appending a string \(v \in \Sigma^+\) with length at most \(l\) to the current state \(s_t\), resulting in the string \(s_t\#v\), to which \(Y\) adds another string \(w \in \Sigma^+\) (\(|w| \le l\)), resulting to \(s_{t+1}=s_t\#v@w\) (``\#'' and ``@'' are \(X\) and \(Y\)'s delimiters, respectively. In reality, they are replaced by speakers designators, such as ``X: ...''). This string is then fed to a string classifier, \(\mc{C}: (\Sigma^*\#\Sigma^+@\Sigma^+)^+ \rightarrow \{0,1,Cont\}\). The classifier's outcome, \(c_t=\mc{C}(s_t)\), determines whether the game needs to stop (if it is \(0\) or \(1\)) or continue to the next stage (if it is \(Cont\), unless a maximum number of stages, \(d\), is reached). In practice, \(Cont\) represents a case that a conclusive verdict cannot be made. In the extended-form formalism, \(\mc{C}(s_t)\) determines if a decision node, \(s_t\), in the game tree is terminal or not. To keep the game general, we propose using separate utility functions, one for each type of the players, in the form of \(u: T_X \times T_Y \times \{0,1,Cont\} \rightarrow \mathtt{R}\). This determines the utility of the players in the terminal nodes.

\begin{definition}
A \emph{verdict game}, is a conversation game parametrised by \((T_X, T_Y, \Sigma, l, d, \mc{C}, \{u_i\}_{i \in T_X \cup T_Y})\) that is played as described in the beginning of this section.
\end{definition}

\begin{example} \label{ex:court-formal}
\textbf{Court (formal).} This is a formalised version of Example \ref{ex:court}. Players \(X\) and \(Y\) are the prosecution and the defence, respectively. The classifier is a judge, where the verdicts of 0 and 1, represent verdicts of ``not guilty'' and ``guilty'', respectively. As long as the players are concerned, the true type of the defence (actually guilty or not) is irrelevant. Therefore, both players are of a single types which are omitted from \(u\) function for brevity. The utility functions are therefore defined as below. \(u_X(Cont)\) reflects the doctrine of \emph{presumption of innocence}.  
\begin{equation}
\begin{aligned}
u_X(0) = -1, u_X(1) = 1, u_X(Count) = -1 \\
u_Y(0) = 1, u_Y(1) = -1, u_Y(Cont) = 1
\end{aligned}
\end{equation}
\end{example}

\begin{example} \label{ex:interrogation-formal}
\textbf{Interrogation (formal).} This is a formalised version of Example \ref{ex:interrogation}. Players \(X\) and \(Y\) are the interrogator and the suspect, respectively. As the suspect could be guilty or non-guilty, \(t_Y=\{\text{Guilty}, \text{Non-Guilty}\}\). At first glance, the classifier might seem to reflect the interrogator's internal judgment about the suspect's status. However, upon closer examination, it becomes clear that the relevant judgment for an effective interrogation is that of an impartial observer—akin to an indifferent judge assessing the interrogation's progression in a courtroom. This perspective is captured by the non-strategic classifier in our model. Again, the verdicts of 0 and 1, represent verdicts of ``guilty'' and ``not guilty'', respectively. However, the goal of the interrogator is to establish the suspect's true type. The respective utility functions are shown in Table \ref{tab:interrogation-formal-utilities}.

\begin{table}[h!]
\centering
\small
\begin{tabular}{ccccc}
\hline
Player & \(t_Y\) & \(c=0\) & \(c=1\) & \(c=Cont\) \\ \hline
\(X\)           & Non-Guilty       & \(1\)            & \(-1\)           & 0                  \\ 
\(X\)           & Guilty           & \(-1\)           & \(1\)            & 0                  \\ 
\(Y\) (Non-Guilty) & Non-Guilty     & \(1\)            & \(-1\)           & \(1\)              \\ 
\(Y\) (Guilty)   & Guilty          & \(1\)            & \(-1\)           & \(1\)              \\ \hline
\end{tabular}
\caption{Utility outcomes for players \(X\) and \(Y\) under different types and actions in an interrogation game.}
\label{tab:interrogation-formal-utilities}
\end{table}

It can be observed that for the interrogator, this is just the usual (unweighted) classification cost function, where false positives and false negatives are penalised while true positives and true negatives are rewarded. Alternative utility functions, that represent various preferences are of course possible. The goal of the suspect is to be exoneration, no matter what his type is.
\end{example}

\begin{example} \label{ex:turing-test-formal}
\textbf{Turing test (formal).} The usual Turing test (Example \ref{ex:turing-test}) is isomorphic to the interrogation example, where the witness types, i.e., ``Non-Guilty'' and ``Guilty'' can correspond to ``Human'' and ``Machine'', respectively. Classifier is an AI-detection system, that decides, based on the history of the conversation at each state, whether the witness is Human (0) or Machine (1).
\end{example}

We conclude this section by putting this approach to the Turing test in context. Historically, during the era of ELIZA \cite{weizenbaum_elizacomputer_1966}, the Turing Test was viewed primarily as a benchmark for intelligence, often involving simple, non-strategic interactions. However, the landscape has changed significantly; advanced Large Language Models (LLMs) are now sometimes judged as ``more human than humans'' \cite{rathi_gpt-4_2024}. Current statistical and machine learning models perform well at detecting AI-generated text \cite{mitchell_detectgpt_2023,mireshghallah_smaller_2024}, but their effectiveness is limited to texts directly produced by LLMs as they are not designed to detect adversarially manipulated texts designed to obscure their machine origins. As deceptive techniques advance, it is likely that more sophisticated detection methods will be necessary—methods that strategically interact with agents to reveal their machine nature, akin to the sequential and interactive \emph{Voight-Kampff} Test from the science-fiction classic \emph{Blade Runner}.

\subsection{Considerations and Limitations}
The Verdict game, despite its versatility, is constrained by several limitations. One major challenge is the enormous branching factor at each step, as it includes all possible utterances that could occur, such as topic changes or tangential remarks. A potential workaround is to narrow down the options to only the most likely and natural utterances for that moment. Alternatively, players can adopt high-level strategies—such as focusing on eliciting contradictions, pressing for details, or establishing inconsistencies—and restrict their utterances to those aligned with their chosen strategy.

Another limitation arises in cases, e.g., where the interrogator seeks specific information, such as the name of a person or non-binary details, which are harder to model within the game's structure. Additionally, the importance of impartial judgment poses a challenge, as the players’ personal judgments might influence outcomes, for example, in scenarios where judges themselves ask questions (true that even here the judge's decision might be judged by a higher-level court or by the public or even the ``history'', but we might equally decide not to delve into this complexity) or when a human suspects that the other side might be a chatbot and it is not easy for them to use AI-detector tools to verify the fact.

\subsection{Equilibria}
All conversation games, including the verdict game, are finite, because the number of stages and the length of player actions are finite. In particular, the court game (Example \ref{ex:court-formal}) is a complete-information, zero-sum game, in principle, solvable by an standard minimax algorithm to a Subgame Perfect Equilibrium \cite[Chapter 8]{tadelis_game_2013}. Admittedly, the action space is too large to allow for an exhaustive search. Therefore, approaches such as Monte Carlo Tree Search \cite{swiechowski_monte_2023} need to be employed to approximately solve the game. The interrogation games, e.g., Examples \ref{ex:interrogation-formal}, \ref{ex:turing-test}, are Bayesian \cite[Chapter 15]{tadelis_game_2013}. The appropriate solution concept is Perfect Bayesian Equilibrium \cite[Chapter 16]{tadelis_game_2013}. Again, the action space is too large, calling for a surrogate methods, such as Information Set Monte Carlo Tree Search \cite{cowling_information_2012}. In the next section, we present the results obtained under a severely limited branching factor as a proof of concept.




\section{Experiments} \label{sec:experiments}
This section serves as a proof of concept illustration of the ideas proposed in this paper, and we leave a through exploration to future work.

\subsection{Court Process}
To simulate the court example (Example \ref{ex:court-formal}), we used LLM agents \cite{huang_understanding_2024,xi_rise_2023}, to act as prosecution, defence, and judge. To illustrate the efficacy of strategic planning, we compared a naive prosecutor with a strategic one. While the naive prosecutor selects questions using the LLM's default temperature, the strategic prosecutor performs a shallow search to introspect and choose the question most likely to convince the judge (Table \ref{tab:court-experiment}). The results of the experiments show that the strategic agent wins \%64 of the times, while the naive agent wins \%27 of the times. The difference is statistically highly significant with a \(p\)-value of \(\le 1e-5\).

\paragraph{Detail of experiments.} LLM: OpenAI GPT-4o \cite{openai_gpt-4o_2024} (\texttt{gpt-4o-2024-11-20}), depth of introspection: 1, breadth of introspection: 10, number of experiments for each type of prosecutor: 100.


\begin{table}[h!]
\setlength{\extrarowheight}{2pt}
\centering
\begin{tabular}{m{2cm}m{5.5cm}}
\hline
Component & Description \\ \hline
Context (Case Details) & \emph{Victim}: Emily Harper (34, journalist) \emph{Suspect}: Ryan Carter (ex-boyfriend) \emph{Evidence}: fingerprints, torn jacket, text messages, being seen by witnesses, signs of struggle in victim's apartment. \\ \hline
\makecell[l]{Play X \\ (Prosecutor, \\ Naive)} & Generates a response based on default temperature settings, asking simple and direct questions without leveraging psychological or strategic patterns. \\
\makecell[l]{Play X \\ (Prosecutor, \\ Naive)} & Introspects using 10 simulated conversations to generate the most effective question. \\ \hline
\makecell[l]{Player Y \\ (Defence)} & Represents Ryan Carter (suspect), responding with plausible denials, evasions, or misleading statements designed to conceal guilt. \\ \hline
Classifier & Analyzes the conversation history and assigns one of three verdicts: Guilty, Innocent, or Non-conclusive, with an emphasis on identifying guilt if vague responses occur to the questins regarding the suspect being seen by witnesses. \\ \hline
\end{tabular}
\caption{Description of components in the court experiment.}
\label{tab:court-experiment}
\end{table}

\section{Related Work} \label{sec:related-works}
This section situates our study within the broader landscape of game theory, linguistic interactions, and AI-related strategic frameworks.

\paragraph{Classic games modelled after linguistic interactions.} Many game models in game theory is modelled after games where the language is a medium for actions, e.g., offering a wage in a signalling game \cite[Chapter 16]{tadelis_game_2013}, and players engaging in \emph{cheap talks} \cite[Chapter 18]{tadelis_game_2013}. However, these are oversimplified models where the actions are discretised into a limited set of predefined choices. In other domain, such as auctions \cite[Chapter 13]{tadelis_game_2013}, the linguistic choices are often irrelevant. This is in contrast with the present work, where the actions as linguistic utterances as such. 

\paragraph{Games over language.} Several studies consider games that involve alphabets. Two such works consider games where players add letters to the end of shared word \cite{rosenfeld_ann_2024,marcus_winning_2023}, with clearly different scope with the present work. Also, there are game theoretic studies of language-based games, such a general study of hidden-role games \cite{carminati_hidden-role_2024}, Werewolf (Mafia) \cite{wang_optimal_2024}, and Secret Hitler \cite{reinhardt_competing_2020}. However, these works either abstract away linguistic interactions or specialise to a specific game \cite{bertolazzi_chatgpts_2023}.

\paragraph{Game theoretic linguistics.} There is a branch of studies that studies pragmatics and rational discourse from a game theoretic standpoint \cite{jacob_regularized_2024,goodman_pragmatic_2016,hintikka_game_2012}. This is in contrast with the present study, where the goal is to define a game using language. 

\paragraph{Strategic classification.} Our verdict game involves strategically taking turns to steer the ongoing conversation toward the boundaries of a sequence classifier. This has similarities with the problem of \emph{strategic classification} \cite{hardt_strategic_2015,ghalme_strategic_2021,ghalme_strategic_2021}, where a player plays against a classifier (i.e., the party that controls the classifier): the player strategically games the classifier by changing its observable attributes to achieve a desirable classification outcome (such as a desirable mortgage decision), while the party that controls the classifier tries to oppose such manipulations. This is in contrast with the present work, where the classifier is non-strategic.

\paragraph{Turing test and AI detection.} One of our motivating examples is the Turing Test \cite{oppy_turing_2021}. At the era of large language models, Several studies has shown significant imporvemnts  advanced Large Language Models (LLMs) are sometimes judged as ``more human than humans.'' \cite{rathi_gpt-4_2024}. Current statistical and machine learning models achieve high accuracy in detecting AI-generated text \cite{mitchell_detectgpt_2023,mireshghallah_smaller_2024}, but these methods are primarily effective for texts directly produced by LLMs and are not designed for texts adversarially manipulated to obscure their machine origins.


\paragraph{LLM agents.} LLM-based agents are AI systems that leverage a large language model (LLM) as their core reasoning engine. A key characteristic of these agents is their ability to engage in multi-turn interactive conversations \cite{huang_understanding_2024,xi_rise_2023}. We utilized LLM-based agents, in our experiments, in dual roles: as players and as judges, each guided by tailored prompts designed for their respective tasks.


\section{Conclusion and Future Work} \label{sec:conclusion}
This paper introduced a formal framework for studying \emph{conversation games}, with a focus on their subset, \emph{verdict games}, where dialogue influences external judgments. By modeling scenarios such as courtroom trials, interrogations, and the Turing test, we showcased how strategic linguistic interactions shape outcomes. Our analysis of the Turing test revealed its strategic depth, emphasizing the importance of robust detection methods in the era of advanced AI systems.

Simulation experiments demonstrated the superiority of strategic agents over naive ones, highlighting the practical relevance of our framework. However, several challenges remain, such as handling the high branching factor in conversational moves and modeling more complex objectives in multi-agent settings.

Future work will focus on exploring trust dynamics in AI-human interactions and developing advanced interrogation strategies. We also aim to leverage advanced search methods, such as Monte Carlo Tree Search, to approximate equilibria in more complex conversation games. This framework provides a foundation for understanding and designing systems that navigate strategic linguistic interactions in both human and AI-driven contexts.

\bibliographystyle{named}
\bibliography{references}

\end{document}